\documentclass[conference]{IEEEtran}
\IEEEoverridecommandlockouts
\usepackage{cite}
\usepackage{multirow}
\usepackage{amsmath,amssymb,amsfonts}
\usepackage{algorithmic}
\usepackage{graphicx}
\usepackage{textcomp}
\usepackage{xcolor}
\def\BibTeX{{\rm B\kern-.05em{\sc i\kern-.025em b}\kern-.08em
    T\kern-.1667em\lower.7ex\hbox{E}\kern-.125emX}}
\begin{document}

\title{Mobile Application for Oral Disease Detection using  Federated Learning\\
}

\author{\IEEEauthorblockN{Shankara Narayanan V}
\IEEEauthorblockA{\textit{Department of Computer Science and Engineering, } \\
\textit{Amrita School of Computing, Coimbatore,}\\
Amrita Vishwa Vidyapeetham, India \\
cb.en.u4cse20656@cb.students.amrita.edu}
\and
\IEEEauthorblockN{Sneha Varsha M}
\IEEEauthorblockA{\textit{Department of Computer Science and Engineering,} \\
\textit{Amrita School of Computing, Coimbatore,}\\
Amrita Vishwa Vidyapeetham, India \\
cb.en.u4cse20659@cb.students.amrita.edu}
\and
\IEEEauthorblockN{Syed Ashfaq Ahmed}
\IEEEauthorblockA{\textit{Department of Computer Science and Engineering,} \\
\textit{Amrita School of Computing, Coimbatore,}\\
Amrita Vishwa Vidyapeetham, India \\
cb.en.u4cse20665@cb.students.amrita.edu}
\and
\IEEEauthorblockN{Guruprakash J}
\IEEEauthorblockA{\textit{Department of Computer Science and Engineering,} \\
\textit{Amrita School of Computing, Coimbatore,}\\
Amrita Vishwa Vidyapeetham, India \\
j\_guruprakash@cb.amrita.edu}
}

\maketitle

\begin{abstract}
The mouth, often regarded as a window to the internal state of the body, plays an important role in reflecting one’s overall health. Poor oral hygiene has far-reaching consequences, contributing to severe conditions like heart disease, cancer, and diabetes, while inadequate care leads to discomfort, pain, and costly treatments. Federated Learning (FL) for object detection can be utilized for this use case due to the sensitivity of the oral image data of the patients. FL ensures data privacy by storing the images used for object detection on the local device and trains the model on the edge. The updated weights are federated to a central server where all the collected weights are updated via The Federated Averaging algorithm. Finally, we have developed a mobile app named OralH which provides user-friendly solutions, allowing people to conduct self-assessments through mouth scans and providing quick oral health insights. Upon detection of the issues, the application alerts the user about potential oral health concerns or diseases and provides details about dental clinics in the user’s locality. Designed as a Progressive Web Application (PWA), the platform ensures ubiquitous access, catering to users across devices for a seamless experience. The application aims to provide state-of-the-art segmentation and detection techniques, leveraging the YOLOv8 object detection model to identify oral hygiene issues and diseases. This study deals with the benefits of leveraging FL in healthcare with promising real-world results. 
\end{abstract}


\section{Introduction}
Oral health is an essential part of a person’s overall health and well-being. It is taught to children from a tender age to follow acts in order to maintain oral health. Oral health involves the healthiness of gums, palate, linings of mouth and throat, tongue, lips, salivary glands, nerves, chewing muscles, and bones of upper and lower jaws\cite{benjamin2010oral}. Studies have indicated that negligence to oral health can result in possible associations between chronic oral infections and diabetes, stroke, heart, and lung disease, and low birth weight or premature births \cite{united2000oral}. So, it is important to monitor our Oral health and practice Oral hygiene. 
Currently, Oral health is monitored predominantly through routine dental examinations where dentists visually inspect the mouth for issues. But this practice of visiting dentists regularly is not followed by a huge number of the population. This is attributed to the fact that potential patients do not have easier access to dentists and dental clinics as they are not distributed evenly \cite{kim2021smart}. Detecting ailments at an earlier juncture holds the potential for enhanced treatment outcomes, increased chances of recovery, or extended periods of survival\cite{lee2004early}.
This creates a requirement for technology that will help people monitor their oral health without much effort and time. There's been a rise in the development of mobile apps and wearable devices designed to help individuals monitor their oral health. These apps might remind users to brush and floss regularly, track their oral hygiene habits, and even provide educational content. However, the development of an application that will scan people’s mouths, detect potential threats, and suggest dentists in the neighborhood will definitely prove to be very useful. This poses a need for deep learning and computer vision techniques that include image segmentation \cite{haralick1985image} and object detection \cite{mack2008object}.
 \begin{figure*}[h]
 \centering
  \includegraphics[width=0.5\textwidth, height=5cm]{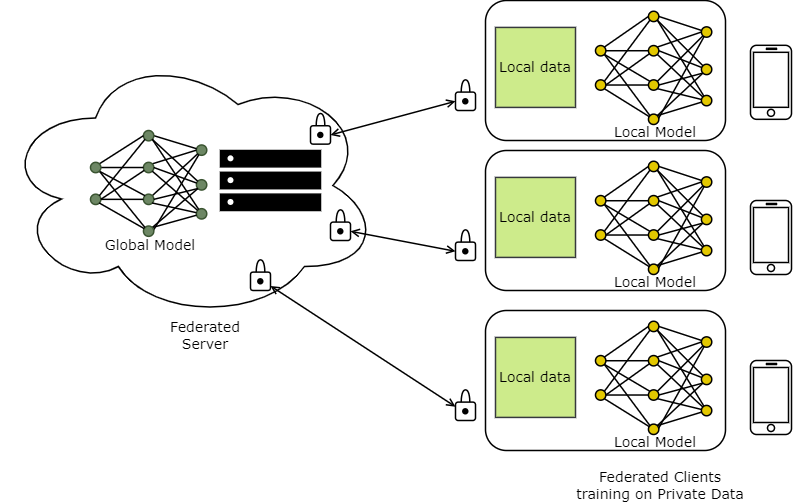}
  \caption{Federated Learning Implementation.}
  \label{fig:Fed_dig}
\end{figure*}
Computer Vision is an important tool used in various industries like the agriculture industry for analyzing grain quality, the security industry for face recognition, the automobile industry for self-driving cars, and even the medical industry for diagnostic imaging \cite{yuen2023machine}. Integration of computer vision with deep learning models can potentially help in diagnosing medical images to know potential threats. However, it is important to tackle the issue of data privacy when using medical data to train these models \cite{bae2018security}. This leads to introducing the concept of federated learning where data remains on the client’s local device and is trained on local models. The weights obtained from training these local models are then sent to the server and aggregated. The aggregated global model is sent back to the clients for future training. This approach preserves data privacy by keeping sensitive information on the client’s side.
The inclusion of a user interface along with these techniques will help in building full-fledged applications that help users examine their oral health and hygiene \cite{yuen2023machine}. By integrating features like dental clinic recommendations and articles on oral health, such applications can offer assistance to people in various ways.

\section{Related Works}
The model discussed by \cite{luo2019real} gives a federated learning approach to a real-world image dataset created by them which consisted of images taken from street cameras and had a total of seven recognizable objects. The paper explores the YOLOv3 and Faster R-CNN models for object detection purposes and compares the performance of the two. The model uses a modified FederatedAveraging (FedAvg) algorithm done by introducing checkpoint saving and restoring on hard devices. This reduces the complication of model aggregation processing.

The model proposed by \cite{roth2020federated} classifies breast density using varied and diverse data with the help of federated learning. The federated algorithm used in this model is Federated Averaging. The study reported an average of 6.3\% better performance against a model trained on a local dataset. Additionally, the model's generalizability was reported to have increased by an average of 45.8\% when evaluated on other unknown testing datasets.
\section{Experiments}
In this section, we will discuss the various experiments conducted, which include disease detection using the object detection frameworks YOLOv5 and YOLOv8, followed by federated learning and lastly constructing a mobile application wherein users will be able to click photos of their mouths and the model would detect diseases using the object detection model which would be modified and improved using federated learning.

\subsection{Dataset Description}\label{AA}
The dental condition dataset is a collection of images curated specifically for dental research and analysis. The dataset comprises a total of 1058 images and four classes encompassing a wide range of dental conditions including Caries, Gingivitis, tooth discoloration, and ulcers. The images are sourced from multiple hospitals and dental websites. This is made to ensure the authenticity and diversity of the dental conditions. 
The dataset is curated using MakeSense AI, which provides a user-friendly interface for annotating and augmenting images. The labeling of data is done in YOLO labeling format with object class references mentioned in Table \ref{tab:instances}.
Furthermore, the dataset is divided into 1465 training images and 43 testing images as displayed in Table \ref{tab:instances}.

\begin{table}[ht]
    \centering
    \caption{Class Counts for training and testing data}
    \begin{tabular}{|c|c|c|c|}
        \hline
        \multirow{2}{*}{Class} & \multirow{2}{*}{Object} & \multicolumn{2}{c|}{Instances} \\
        \cline{3-4}
        & & Training & Testing \\
        \hline
        0 & Caries & 2051 & 35\\
        \hline
        1 & Ulcer & 364 & 5\\
        \hline
        2 & Tooth Discoloration & 315 & 18\\
        \hline
        3 & Gingivitis & 653 & 106\\
        \hline
    \end{tabular}
    \label{tab:instances}
\end{table}

 \begin{figure*}[h]
 \centering
  \includegraphics[width=0.8\textwidth, height=6cm]{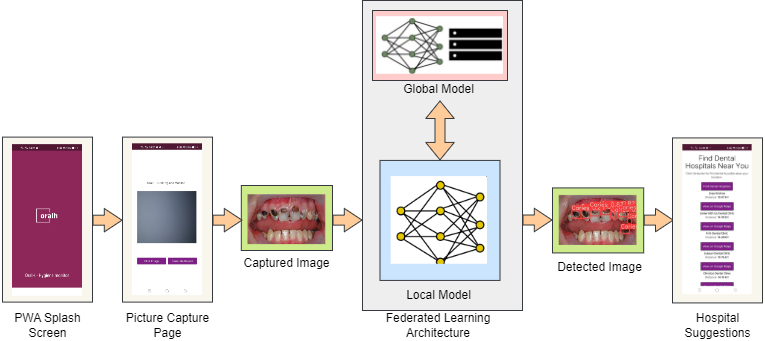}
  \caption{Oral Disease Detection Pipeline}
  \label{fig:Flow_dig}
\end{figure*}

\subsection{Object Detection}
The main idea behind object detection is to localize a region of interest and giving a class to this region like an image classifier. The presence of many regions of interest makes it an advanced problem of image classification. YOLO (You Only Look Once) \cite{redmon2016you} is a model significantly known for its object detection accuracy and speed, and is often used for multiple object detection \cite{10084220}. It uses an end-to-end neural network that makes predictions of class probabilities and bounding boxes at once. YOLO takes an input image and divides it into a P*P grid. Given the grid cell contains the center of an object, that particular grid cell is responsible for detecting that object. Every grid cell has the possibility to predict Q bounding boxes and can give confidence scores for the boundary boxes. A post-processing step called Non-Maximum Suppression (NMS) \cite{redmon2016you} is incorporated in YOLO to improve the efficiency of object detection. It is used to distinguish and eliminate incorrect and redundant bounding boxes to output a single bounding box for each image object. The YOLO model's performance in the detection of real-world data is the primary reason for selecting this object detection framework over other existing models \cite{9215321}.
YOLOv5 \cite{jocher2022ultralytics} and YOLOv8 \cite{jocher2023yolo} models are used in this application to perform object detection. YOLO models which do single-shot object detection are better compared to two-shot object detection models like Faster RCNN \cite{ren2015faster} for this particular application. In general, single-shot object detection is suited better for real-time applications when compared to two-shot object detection models. As the concept of federated learning is also incorporated, it is important to perform object detection faster in order to avoid latency. Hence, the single-shot object detection model, YOLO, was chosen for this particular use-case due to its suitability when compared to RCNN-based models that follow two-shot object detection.

\subsection{Federated Learning}
Federated Learning is an innovative framework in the field of machine learning and artificial intelligence that proposes a mechanism to tackle problems in artificial intelligence such as data privacy and scalability. Groundbreaking work has been achieved in recent times to produce relevant and necessary advancements in this field that have profound impacts on society such as \cite{10066322} which uses federated learning practices to predict the spread of the COVID-19 pandemic in India and \cite{9580097} which uses autonomous vehicle data to predict steering angle and discusses the impact of the communication a vehicle has with its surroundings.

The model we propose uses a client-server-based federated learning approach, wherein the model is trained locally on the data collected by each client. Upon completion of a round of local training, the modified model weights are forwarded to the server for aggregation. Following aggregation, the updated weights are distributed to the clients to start the next round of training. Models at each client converge after the completion of a certain number of FL rounds, where the client chooses the locally best model based on a performance metric. The best model will be chosen by comparing the performance metric of the global model sent by the server and the intermediate local models trained on the client side. The federated averaging algorithm as proposed by \cite{mcmahan2017communication} was used for model weights upgradation. The three primary parameters of the FedAvg algorithm that affect the performance are:
\begin{itemize} 
\item Number of Clients (C)
\item Number of tries performed on the local dataset by each client during a round (E)
\item Minibatch size used for client updates (B)
\end{itemize}
Throughout the experiment, the number of clients(C) used was 5 and the minibatch size(B) was maintained at 1. As illustrated in Fig. \ref{fig:Fed_dig}  the server communicates with the clients and uses local weights of the client to update the central model using the federated averaging algorithm.

\begin{equation}
f(w) = \sum_{k=1}^{K} \left(\frac{n_k}{n}\right) \cdot F_k(w)
\label{eq:fed_avg}
\end{equation}

The Eq. \ref{eq:fed_avg} provides the mathematical description of the federated averaging algorithm, where $F_k (w)$ refers to the loss function of each client, $k$ refers to the number of clients, and $\frac{n_k}{n}$ refers to the weighted average based on the size of the clients' dataset.

\subsection{Mobile Application}
The designed model is combined with the user interface to build an application that helps people monitor their oral health. The application is made as a Progressive Web Application(PWA) instead of a native mobile application. The reason why PWAs were chosen over native mobile applications is that developing PWAs is more cost-effective when compared to native mobile apps since they consolidate all development efforts into web apps rather than requiring the creation of separate applications for each mobile platform \cite{fortunato2018progressive}. The web application was built with HTML, CSS, and JavaScript for the front end, while the back end was built using PHP and MySQL. The PWA features are incorporated by adding a web manifest file and icons. To enable offline functionality and caching, service workers were implemented.

The camera module is integrated using the WebRTC API \cite{7160422}, and it ensures that the flashlight is kept on when a user is using the app on a mobile platform. This ensures proper image capture without leaving out any details, enabling better analysis. It is seamlessly integrated with geolocation services \cite{5672223} to offer personalized recommendations of nearby hospitals and dental clinics in case of detection of potential problems to prioritize prompt professional care. An SQL database containing oral health articles is created and is suggested to the user based on their oral health assessment results.

Figure \ref{fig:Flow_dig} explains the pipeline of oral disease detection using the YOLOv8 model through a federated learning approach. 

\section{Results and Discussions}
This section discusses the performance of the object detection models YOLOv5 and YOLOv8 when trained using local data alone and when federated learning is used. The performance metrics used in evaluating the two frameworks are:
\begin{itemize} 
\item Mean Average Precision
\item F1-Score
\end{itemize}

Given below are the mathematical representations of the evaluation metrics:
\begin{equation}
    mAP = \frac{1}{n} \sum_{k=1}^{n} AP_k \label{eq:mAP}
\end{equation}

\begin{equation}
F1\text{-}Score = 2 \cdot \frac{Precison \cdot Recall}{Precision+Recall} \label{eq:f1}  
\end{equation}

\begin{table}[ht]
    \centering
    \caption{Performance Metrics of Federated Framework \\ Number of Communication rounds required to achieve the target mAP \\ Target mAP = 80\%}
    \label{tab:perf_metr_fed}
    \begin{tabular}{|c|c|c|c|}
        \hline
        Model & E & Rounds \\
        \hline
        Federated - YOLOv5 & 1 & 168 \\
        Federated - YOLOv5 & 5 & 59 \\
        Federated - YOLOv5 & 10 & 122 \\
        Federated - YOLOv8 & 1 & 64 \\
        Federated - YOLOv8 & 5 & 51 \\
        Federated - YOLOv8 & 10 & 97 \\
        \hline
    \end{tabular}
\end{table}
\begin{table}[ht]
    \centering
    \caption{Performance Metrics of Local Training}
    \label{tab:perf_metr_loc}
    \begin{tabular}{|c|c|c|}
        \hline
        Model & F1-Score & mAP \\
        \hline
        YOLOv5 & 75.7\% & 72.6\% \\
        YOLOv8 & \textbf{82.3\%} &\textbf{ 78.8\%} \\
        \hline
    \end{tabular}
\end{table}

As evident from Tables \ref{tab:perf_metr_fed} and \ref{tab:perf_metr_loc}, and discussed in \cite{luo2019real}, initially a larger E value gave faster rates of achieving the target mAP. But as the dataset is more non-IID, different values of E produce different rates of convergence.

The YOLOv8 model contains significant improvements over its preceding YOLOv5 model. The YOLOv8 model utilizes an anchor-free model to predict the center of an object without using an offset from the anchor box. Additionally, the YOLOv8 model shows notable enhancement in accuracy when tested on the COCO dataset \cite{solawetz2023yolov8}. On comparing the two YOLO models on locally trained data, it is evident that the YOLOv8 model performs better than the YOLOv5 model as tabulated in Table \ref{tab:perf_metr_loc}. 

\begin{figure}[h]
    \centering
    \includegraphics[width=0.7\linewidth]{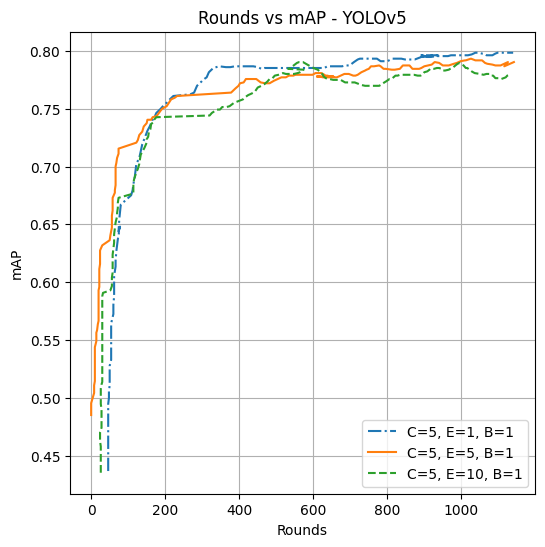}
    \caption{mAP vs. Number of Communication Rounds for the YOLOv5 Model}
    \label{fig:yolo5_mAP}
\end{figure}

Mean Average Precision(mAP) is an evaluation metric often used to evaluate the performance of an object detection model. It utilizes the concept of Intersection over Union(IoU) to measure the overlap of the predicted object boundary against the ground truth boundary. A threshold of 0.5 was used for IoU during the experiments. The precision obtained from this is calculated for all objects and grouped and averaged on the basis of the class to which they belong.

\begin{figure}[h]
    \centering
    \includegraphics[width=0.7\linewidth]{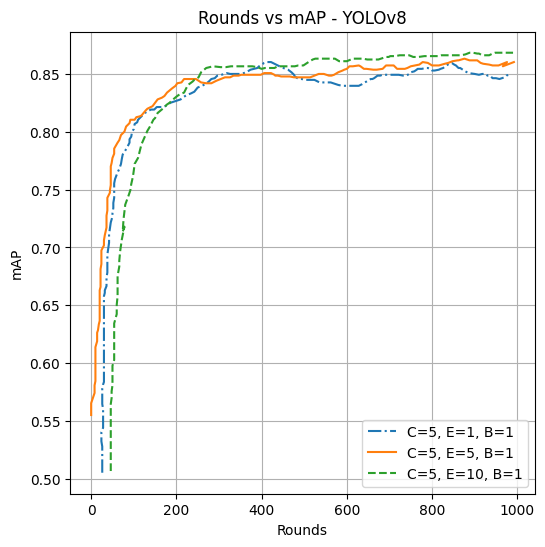}
    \caption{mAP vs. Number of Communication Rounds for the YOLOv8 Model}
    \label{fig:yolo8_mAP}
\end{figure}

Calculation of the mean of the average precision values provides the final mean average precision metric. Eq. \ref{eq:mAP} provides the mathematical description of the metric where AP refers to the average precision of each class and n refers to the total number of classes. 
The locally trained framework produced an mAP value of 72.6 after the completion of 100 epochs on the YOLOv5 model and a value of 78.8 after the completion of 100 epochs on the YOLOv8 model while the federated learning framework achieved a target mAP of 80 after a certain number of communication rounds with the server depending on the value of E and the model in use as evident from Tables \ref{tab:perf_metr_fed} and \ref{tab:perf_metr_loc} as well as Figures \ref{fig:yolo5_mAP} and \ref{fig:yolo8_mAP}. This indicates the overall better performance of the federated framework.

The F1-score is a machine learning performance metric that utilizes the precision and recall metrics. Popularly used for object detection in imbalanced datasets, it provides an integration of the precision and recall metrics and is hence used as an ideal measure of performance for object detection purposes. Eq. \ref{eq:f1} provides the mathematical description of F1-Score.

As evident from Table \ref{tab:perf_metr_loc} the YOLOv8 model provides a better F1-score in comparison to the YOLOv5 model and is hence considered an overall better model for implementation of object detection for this use-case.
\section{Conclusions and Future Work}
In conclusion, this study was able to successfully detect the diseases in the mouth using the YOLOv5 and YOLOv8 models trained locally as well as trained iteratively via Federated Learning(FL). As evident from the results discussed, FL was able to achieve better results for both the YOLOv5 and the YOLOv8 models in comparison to its counterparts. The use of varied and diverse data from different end users provides a better training opportunity for the model and hence the expected results. Between the two YOLO models, the YOLOv8 gave better performance and was hence used for the final application. Amongst the federated models, we noticed that an increase in the value of E does not necessarily imply a faster rate of convergence to achieve the target mAP, as also noted in \cite{luo2019real}.  

Future research direction includes the incorporation of explainable AI into federated learning \cite{barcena2022fed}. Innovative methods such as federated model inspection can be incorporated into federated learning to provide greater insights into the working and decision-making aspects across the devices to the stakeholders involved. Privacy-preserving methodologies while guaranteeing interpretability are essential.

The federated learning framework involves the communication between the server and the clients. This raises the need for secure network and communication protocols to prevent data interception. Network security aspects such as encryption, server authentication, data integrity, and secure channels must be incorporated to protect sensitive data, especially in the case of healthcare applications. Preserving data integrity ensures trust in the system and allows the framework to collaborate with decentralized edge devices.

\bibliographystyle{IEEEtran}
\bibliography{references}

\end{document}